\PassOptionsToPackage{table}{xcolor}

\documentclass[letterpaper, 10 pt, conference]{ieeeconf}  

\IEEEoverridecommandlockouts                              

\usepackage{anyfontsize}
\usepackage{ stmaryrd }
\usepackage{amsmath,amsfonts,bm}
\usepackage{cuted}
\usepackage[ruled,vlined,linesnumbered]{algorithm2e}
\usepackage{graphicx}
\usepackage{tabularx} 
\usepackage{amssymb}
\usepackage{multirow}

\usepackage{floatrow}
\usepackage{color}
\usepackage[bb=boondox]{mathalfa}
\usepackage{dsfont}
\usepackage{upgreek}
\usepackage{mathrsfs}
\usepackage{url}

\usepackage{pifont}
\def\eqref#1{equation~\ref{#1}}

\def\1{\bm{1}}

\DeclareMathAlphabet{\mathsfit}{\encodingdefault}{\sfdefault}{m}{sl}
\SetMathAlphabet{\mathsfit}{bold}{\encodingdefault}{\sfdefault}{bx}{n}

\usepackage{color}
\usepackage[table,xcdraw]{xcolor}
\usepackage{amsmath}
\usepackage{amssymb}
\usepackage{latexsym}
\usepackage{url}
\usepackage{cite}
\usepackage{relsize}
\usepackage{multirow}
\usepackage{afterpage}
\usepackage{ifthen}
\usepackage{graphicx}
\usepackage{float}

\restylefloat{figure}
\usepackage{algorithm2e}
\usepackage{subfigure}
\usepackage{subcaption}
\usepackage{adjustbox}
\usepackage{epstopdf}
\usepackage{stackengine}
\usepackage{gensymb}
\usepackage{booktabs}
\usepackage{makecell}
\usepackage{hhline}
\usepackage{lipsum}
\usepackage{ upgreek }
\usepackage{mathtools}

\usepackage[bb=boondox]{mathalfa}

\usepackage{{flushend}}

\usepackage{tikz}
\usepackage{comment}
\usepackage{wrapfig}
\usepackage[normalem]{ulem}

\definecolor{cite_color}{RGB}{240,150,12}
\makeatletter
\let\NAT@parse\undefined
\makeatother
\usepackage[bookmarks=false, linkcolor=blue, urlcolor=blue, citecolor=blue]{hyperref} 

\hypersetup{
    colorlinks=true,
    linkcolor=red,
    filecolor=magenta,      
    urlcolor=blue,
    pdfstartview={FitH},
    citecolor=blue
    }

\title{\LARGE \bf
FedEFM: Federated Endovascular Foundation Model with Unseen Data\\ 
}

\author{Tuong Do$^{1,2}$, Nghia Vu$^{2}$, Tudor Jianu$^{1}$, Baoru Huang$^{1}$, Minh Vu$^{3}$, Jionglong Su$^{4}$, Erman Tjiputra$^{2}$, \\ Quang D. Tran$^{2}$, Te-Chuan Chiu$^{5}$, Anh Nguyen$^{1}$
\thanks{$^{1}$Deparment of Computer Science, University of Liverpool, UK
        }%
\thanks{$^{2}$AIOZ Ltd., Singapore}%
\thanks{$^{3}$Automation \& Control Institute, TU Wien, Austria}%
\thanks{$^{4}$Xi'an Jiaotong-Liverpool University, China}%
\thanks{$^{5}$National Tsing Hua University, Taiwan}%
}

\begin{document}

\maketitle
\thispagestyle{empty}
\pagestyle{empty}



\begin{abstract}
In endovascular surgery, the precise identification of catheters and guidewires in X-ray images is essential for reducing intervention risks. However, accurately segmenting catheter and guidewire structures is challenging due to the limited availability of labeled data. Foundation models offer a promising solution by enabling the collection of similar-domain data to train models whose weights can be fine-tuned for downstream tasks. Nonetheless, large-scale data collection for training is constrained by the necessity of maintaining patient privacy. This paper proposes a new method to train a foundation model in a decentralized federated learning setting for endovascular intervention. To ensure the feasibility of the training, we tackle the unseen data issue using differentiable Earth Mover's Distance within a knowledge distillation framework. Once trained, our foundation model's weights provide valuable initialization for downstream tasks, thereby enhancing task-specific performance. Intensive experiments show that our approach achieves new state-of-the-art results, contributing to advancements in endovascular intervention and robotic-assisted endovascular surgery, while addressing the critical issue of data sharing in the medical domain.


\end{abstract}
\section{Introduction}
Endovascular surgery is now usually a minimally invasive procedure that diagnoses and treats vascular diseases with several advantages such as reduced trauma and quick recovery time~\cite{ramadani2022survey}. During endovascular surgery, surgeons use catheters and guidewires to access arteries. However, this procedure also entails risks such as potential vessel wall damage~\cite{jianu2022cathsim}. Precise identification of catheters and guidewires within X-ray images is crucial for patient safety~\cite{pereira2022use}. The rise of deep learning has played a vital role in improving surgical precision and enhancing patient safety in endovascular intervention~\cite{huang2023detecting}. However, accurately segmenting intricate catheters and guidewires in X-ray images remains challenging due to the limited quantity of data~\cite{ramadani2022survey}.

Recently, vision language models have received attention from researchers from various domains~\cite{wang2023foundation, zhang2023text}. For example, CLIP~\cite{radford2021learningCLIP} and ALIGN~\cite{jia2021scalingALIGN} demonstrate proficiency in cross-modal alignment and zero-shot learning tasks. 
In the medical domain, EndoFM~\cite{wang2023foundation} is developed as a foundation model for endoscopy video analysis. The LVM-Med model~\cite{nguyen2023lvm} is introduced as a foundation model for medical images across multiple modalities. Although these models show promising results on downstream tasks, \textit{most assume that the data can be collected and trained centrally}, which is usually challenging in the medical domain.

\begin{figure}[!t]
  \centering
\setlength{\tabcolsep}{3pt}
\begin{tabular}{ccc}
\shortstack{\includegraphics[width=0.41\linewidth, height=0.41\linewidth]{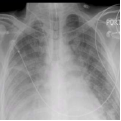}\\\footnotesize (a) Human X-ray}&
\shortstack{\includegraphics[width=0.41\linewidth, height=0.41\linewidth]{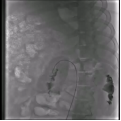}\\ \footnotesize (b) Animal X-ray}\\
\shortstack{\includegraphics[width=0.41\linewidth, height=0.41\linewidth]{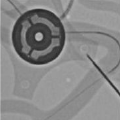}\\\footnotesize (c) Phantom X-ray}&
\shortstack{\includegraphics[width=0.41\linewidth, height=0.41\linewidth]{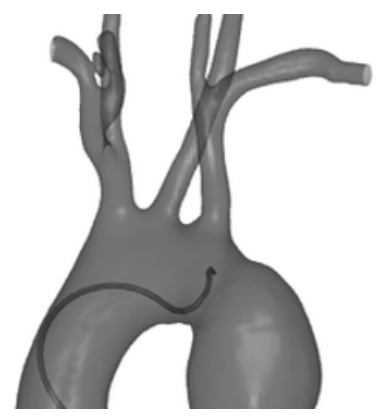}\\ \footnotesize (d) Simulation X-ray}\\
\end{tabular}
\vspace{-4ex}
    \caption{Different types of endovascular X-ray data. \vspace{-5ex}}
    \label{fig:intro}
\end{figure}

In practice, collecting large-scale data in the medical domain is not a trivial task due to data privacy~\cite{kaissis2020secure,huang2022simultaneous}. To overcome this limitation, federated learning is emerging as a candidate, enabling the training process to occur between hospital silos without collecting patient data. 
Despite the advantages of federated learning, current challenges include ensuring convergent training across different silos~\cite{jiang2022harmofl} and heterogeneous data~\cite{zhou2023fedcontrast}. In endovascular intervention, these challenges primarily stem from data gathered from various sources, hence leading to the domain gap between X-ray data. Fig.~\ref{fig:intro} shows an example of X-ray images from different endovascular datasets. We observe that due to privacy, endovascular datasets with real human X-ray images are usually small, compared to data collected with animal, silicon phantom models, or from simulation environments~\cite{huang2024cathaction}.

In this paper, our goal is to train a foundation model using diverse endovascular datasets with federated learning. Since we aim to use all possible endovascular data (i.e., from humans, animals, phantoms, etc.), there is an unseen data problem between silos (Fig.~\ref{fig:UnseenIssue}). To tackle this problem, we propose the Federated Endovascular Foundation Model (FedEFM), a new distillation algorithm using differentiable Earth Mover's Distance (EMD). Once trained, FedEFM provides crucial initializations for downstream tasks, thereby enhancing task-specific performance. Our approach outperforms existing methods and holds significant potential for application in robotic-assisted endovascular surgery, while effectively maintaining data privacy.

Our contribution can be summarized as  below:

\begin{itemize}
    \item We propose a new method to train a federated endovascular foundation model with unseen data using a multishot distillation technique.
   \item We collect new datasets for training endovascular foundation models. Our proposed model is verified under several downstream tasks. Our code will be released.
\end{itemize}
\section{Literature Review}
\label{Sec:Literature}

\textbf{Endovascular Intervention.}
Endovascular intervention has significantly advanced the treatment of various vascular diseases, such as aneurysms and embolisms under X-ray fluoroscopy~\cite{baert2003guide, zhou2020real, gherardini2020catheter}. However, these procedures face several challenges due to poor contrast~\cite{moore2005mri}, the complexity of anatomical structures~\cite{efthymiou2023factors}, and the limited availability of expert-labeled data~\cite{breininger2018multiple, breininger2018intraoperative}. Recent research has focused on improving these aspects through advanced imaging technologies and machine learning approaches~\cite{ma2021tensor, ranne2023aiareseg, mei2023real }. Specifically, the authors in in~\cite{du2024guidewire} proposed an improved U-Net-based method for guidewire endpoint localization in X-ray images. 
Recently, FW-Net~\cite{nguyen2020end} is proposed to enhance catheter segmentation by leveraging frame-to-frame temporal continuity. While several works focus on traditional tasks, few develop foundation models for endovascular intervention~\cite{ nguyen2023lvm,wang2023foundation}. The main reason is that patient data must be kept private, which becomes a major barrier preventing foundation models from being trained~\cite{wang2022medclip, MedSAM, koleilat2024medclip}. 

\textbf{Federated Learning.}
Federated learning has emerged as a promising solution for training machine learning models on decentralized data while preserving data privacy~\cite{mammen2021federated}. This approach is particularly beneficial in medical domains, where data sensitivity and privacy concerns are paramount~\cite{jiang2023client}. Although various studies have explored the application of federated learning to train foundation models in the medical field~\cite{chen2024feddat, liu2024fedfms}, the privacy issue can be handled but not fully resolved~\cite{xue2023differentially, okegbile2023differentially }. The inherent heterogeneity and non-IID nature of medical data across different institutions present significant challenges~\cite{lin2023heterogeneous, wang2024towards}. Additionally, the unseen data issue, where certain types of data are present in some datasets but absent in others, complicates the training process and model generalization~\cite{liu2021feddg, tolle2024funavg}.


\textbf{Knowledge Distillation with Earth Mover's Distance.}
Knowledge distillation involves transferring knowledge from a large, complex model (the teacher) to a smaller, more efficient model (the student)~\cite{hinton2015distilling}. In the context of federated learning, distillation can be used to enable local models to learn from aggregated global models without sharing raw data\cite{jeong2018communication}. The Earth Mover's Distance (EMD), also known as the Wasserstein distance, measures the dissimilarity between two probability distributions and is particularly useful for comparing distributions that do not have overlapping support. By leveraging the differentiable EMD, it is possible to align distributions of labels across different models, facilitating better model convergence and knowledge transfer~\cite{zhang2020deepemd, zhao2008differential}. In this paper, we leverage EMD within a distillation training process to address the unseen label data issue when training endovascular foundation models in federated scenarios.

\section{Federated Endovascular Foundation Model}
\label{sec:preliminary}

\subsection{Motivation}

\begin{figure}[!h]
  \centering
  \includegraphics[width=1.0\linewidth]{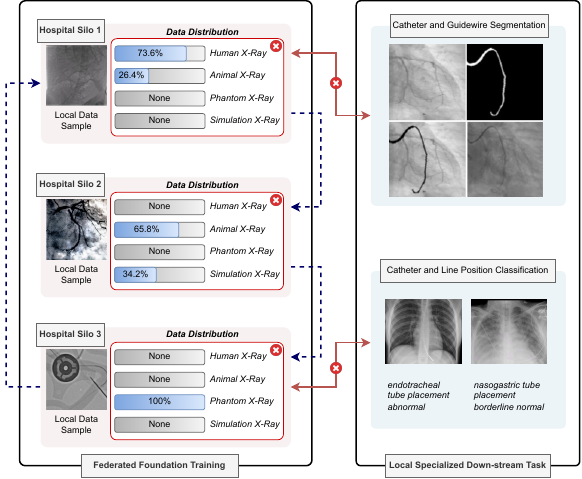}
 \caption{Unseen data issue visualization. Red lines with crosses indicate insufficient data for training. Blue dotted lines between data silos indicate transferable weights. 
 }
 \label{fig:UnseenIssue}
\end{figure}

We aim to train a federated foundation model for endovascular intervention with all possible types of X-ray data. In practice, each silo (hospital) retains certain data sources that may not be available at other hospitals. The issue arises from the dissimilarity in data corpora across hospitals, i.e., some data are available in one hospital but not in others. Fig.~\ref{fig:UnseenIssue} shows an illustration of this problem. Consequently, this leads to the \textit{unseen data} issue that needs to be addressed to ensure the feasibility of the federated training process. 

According to ~\cite{marfoq2020throughput}, while federated learning upholds data privacy by prohibiting direct data sharing, it permits the transmission of model weights between connected hospital silos. To take advantage of this characteristic, we propose the Federated Endovascular Foundation Model (FedEFM), a multishot foundation federated distillation algorithm using EMD to ensure the feasibility of learning. Specifically, our approach enables a local silo model to learn from its neighbors' data and subsequently integrate the acquired knowledge back into the original silo through a distillation mechanism. Unlike other approaches that require a similar label set in both local and global models trained on contributed silos, devices, or servers~\cite{do2023addressingCDL,marfoq2020throughput,nguyen2020autonomous}, our method allows a smooth federated training procedure where hospitals do not need to share their data corpora, thus further improving data privacy. Moreover, once trained, the foundational model's weights serve as valuable initialization for downstream tasks.

\subsection{Federated Distillation with EMD}




We propose Algorithm~\ref{alg:KL} for training a foundation model within a decentralized federated learning process, effectively addressing the issue of the unseen data problem. 
Specifically, in the initial round, local model weights  $\theta_i$ of each $i$-th hospital silo is trained using their respective local data $\xi_i$. Note that $N$ is the maximum number of hospital silos. Within the next communication round, we first perform overseas training where local model weights $\theta_i$ of each $i$-th silo is transmitted to each of their $j$-th neighbor hospital silo; $j \in \mathcal{N}(i)$ denotes a list of neighbors of $i$-th silo. This process aims to let local weights $\theta_i$ learn knowledge from the data $\xi_j$ of its $j$-th neighbor silo. 

We consider $\theta_{i\rightarrow j}$, the so-called overseas expert, to denote the weight of silo $i$ being transmitted to silo $j$ to learn external knowledge. In $(k+1)$ specific communication round, each transferred weight $\theta_{i\rightarrow j}$ is optimized in $j$-th silo using the Equation~\ref{eq:intersilo}.
\begin{equation}
\begin{aligned}
\theta_{i\rightarrow j}(k+1)=   \theta_{i\rightarrow j}\left(k\right)-\alpha_{k}\nabla  \mathcal{L}_{c}\left(\theta_{i\rightarrow j}\left(k\right),\xi_j\left(k\right)\right)  
\end{aligned}
\label{eq:intersilo}
\end{equation}
where $\xi$ is the data in a mini-batch, $\alpha$ is the learning rate, and $\mathcal{L}_{c}$ is the Cross-Entropy loss used for training a typical foundation classification model~\cite{hinton1995wakeCrossEntropy}.



\begin{algorithm}[!ht]
\caption{Federated knowledge distillation with Earth Mover’s Distance.
\label{alg:KL}}
\SetKwInput{KwInput}{Input}                
\SetKwInput{KwOutput}{Output}              
\DontPrintSemicolon
  \KwInput{Initial weight $\theta_i(0)$ for each silo $i$; Maximum training round $K$. }
     
     \For{$k = 0$ \KwTo $K - 1$}{
     
     
     \tcp{The loop below is parallel}
     \ForEach{$\text{silo } i$}{
     $\mathcal{N}(i)\leftarrow$ List of $i$-th neighbour nodes.

     $\xi_i\left(k\right) \leftarrow$ Sampling data from local silo $i$
     
    
     \ForEach{$\text{silo } j \in \mathcal{N}(i)$}{
      $\xi_j\left(k\right) \leftarrow$ Sampling data from the $j$-th neighbor of silo $i$
      
      $\theta_{i \rightarrow j}\leftarrow$ Train overseas expert model at $j$-th silo using Equation~\ref{eq:intersilo}.

      $\hat \theta_{i \rightarrow j} \leftarrow \theta_{i \rightarrow j}$  \tcp{Collect overseas expert weights from $j$-th neighbor back to $i$-th silo.}

      $\text{EMD}(\theta_{i}, \hat \theta_{i \rightarrow j}) )\leftarrow$  Compute Earth Mover's Distance using Equation~\ref{eq:EMD}.
      
      }
      
      $\theta_i (k+1)\leftarrow$ Compute $\mathcal{L}^i_{\rm MD}$ with Equation~\ref{eq:distil_loss} and train $i$-th local model using Equation~\ref{eq:cross_learn}.
     }
     }

\end{algorithm}

Then, we perform knowledge transfer where each learned overseas expert $\theta_{i\rightarrow j}$ from the previous step is transferred back to the $i$-th silo.
Successfully transferred weights is denoted as $\hat \theta_{i\rightarrow j}$ which shares values with $\theta_{i\rightarrow j}$.


In the local silo $i$, the local weight is updated based on both the original weight $\theta_{i}$ and the transferred weights $\hat \theta_{i\rightarrow j}$ that is learned from the neighbour silo $j$. In particular, we aim to find regions that share similarities between two weights using the Earth Mover’s Distance $\text{EMD}( \theta_{i}, \hat \theta_{i\rightarrow j})$. In this way, the distance measures the contribution of transferred weights during distillation, enabling the local silo to learn from its neighbors while avoiding divergence when weight convergence goals differ significantly. Local weights $\theta_{i}$ is then optimized using:
\begin{equation}
\footnotesize
\begin{aligned}
&\theta_{i}(k+1) =  \theta_i(k)-\\
&   \alpha_{k}\sum_{j \in \mathcal{N}(i)}\text{EMD}( \theta_{i}, \hat \theta_{i\rightarrow j},k)\nabla  \mathcal{L}^i_{\rm MD}\left({\theta}_i\left(k\right),{\hat \theta}_{i\rightarrow j}\left(k\right),\xi_i\left(k\right)\right)
\end{aligned}
\label{eq:cross_learn}
\end{equation}
where $\mathcal{L}_{\rm MD}$ is the distillation loss (Equation~\ref{eq:distil_loss}), and $\mathcal{N}(i)$ indicates in-neighbors of silo $i$.

\textbf{Differentiable Earth Mover’s Distance.}
Assume that the input sample $\xi_i$ from $i$-th local silo passes through the foundation architecture $\theta_{i}$ to generate the dense representation $\mathbf{U} \in \mathbb{R}^{H \times W \times C}$, where $H$ and $W$ denote the spatial size of the feature map and $C$ is the feature dimension. In a parallel manner, $\mathbf{V} \in \mathbb{R}^{H \times W \times C}$ also denotes the dense representation when $\xi_i$ passes through $\hat{\theta}_{i\rightarrow j}$.

Under Earth Mover circumstance, $\mathbf{V}$ represents suppliers transporting goods to demanders $\mathbf{U}$. Then, $\text{EMD}$ between two feature sets $\mathbf{U} = \{u_1, u_2, \ldots, u_{HW}\}$ and $\mathbf{V} = \{v_1, v_2, \ldots, v_{HW}\}$ can be computed as:
\begin{equation}
\small
\begin{aligned}
\text{EMD}(\theta_{i}, \hat \theta_{i \rightarrow j}) = \text{EMD}(\mathbf{U}, \mathbf{V}) = \sum_{p=1}^{HW} \sum_{q=1}^{HW} (1 - c_{pq}) \tilde{x}_{pq}.
\label{eq:EMD}
\end{aligned}
\end{equation}
where $\tilde{x}$ is conducted from optimal matching flow $\tilde{X} = \{x_1, x_2, \ldots, x_{pq}\} $ for each sample pair of two sets $\mathbf{U}$ and $\mathbf{V}$; $c_{pq}$ is the cost per unit transported from supplier to demander and is obtained by computing the pairwise distance between embedding nodes $u_p \subset \mathbf{U}$ and $v_q \subset \mathbf{V}$.

The cost per unit $c_{pq}$ is computed as below and also plays a virtual role in computing the optimal matching flow:
\begin{equation}
\begin{aligned}
c_{pq} = 1 - \frac{u_p^T v_q}{\|u_p\|\|v_q\|}
\label{eq:cost}
\end{aligned}
\end{equation}
where nodes with similar representations tend to generate small matching costs between each other. 
Then, the optimal matching flow $\tilde{X}$ is conducted by optimizing $\tilde{x}$ as below: 
\begin{equation}
\small
\begin{aligned}
\underset{x}{\text{minimize}} \quad & \sum_{p=1}^{HW} \sum_{q=1}^{HW} c_{pq} x_{pq} \\
\text{subject to} \quad & x_{pq} > 0, \quad p = 1, \ldots, HW, \; q = 1, \ldots, HW\\
& \sum_{p=1}^{HW} x_{pq} = v_q, \quad q = 1, \ldots, HW \\
& \sum_{q=1}^{HW} x_{pq} = u_p, \quad p = 1, \ldots, HW
\end{aligned}
\label{eq:EMDopt}
\end{equation}

Here, EMD seeks an optimal matching $\tilde{X}$ between suppliers and demanders such that the overall matching cost is minimized. The global optimal matching flows $\tilde{X}$ can be achieved by solving a Linear Programming problem (LP). 
For the sake of completeness, we transform the optimization in Equation~\ref{eq:EMDopt} to a compact matrix form:
\begin{equation}
\small
\begin{aligned}
\underset{x}{\text{minimize}} \quad & c(\theta)^T x \\
\text{subject to} \quad & G(\theta)x \leq h(\theta),\\
& A(\theta)x = b(\theta).
\label{eq:compactOpt}
\end{aligned}
\end{equation}

Here $x \in \mathbb{R}^{HW \times HW}$ is our optimization variable.
$Ax = b$ represents the equality constraint and $Gx \leq h$ denotes the inequality constraint in Equation~\ref{eq:EMDopt}. Accordingly, the Lagrangian of the LP problem in Equation~\ref{eq:compactOpt} is given by:
\begin{equation}
L(\theta, x, \nu, \lambda) = c^T x + \lambda^T (Gx - h) + \nu^T (Ax - b),
\label{eq:LP}
\end{equation}
where $\nu$ denotes the dual variables on the equality constraints and $\lambda \geq 0$ denotes the dual variables on the inequality constraints.
Following the KKT conditions, we obtain the optimum $(\tilde{x}, \tilde{\nu}, \tilde{\lambda})$ of the objective function by solving $g(\theta, \tilde{x}, \tilde{\nu}, \tilde{\lambda}) = 0$ with primal-dual interior point methods, where
\begin{equation}
g(\theta, x, \nu, \lambda) = \begin{bmatrix}
\nabla_{\theta} L(\theta, x, \nu, \lambda) \\
\textbf{diag}(\lambda)(G(\theta)x - h(\theta)) \\
A(\theta)x - b(\theta)
\end{bmatrix}.
\label{eq:approxKKT}
\end{equation}

Then, with the theorem below, we can derive the gradients of the LP parameters.

Suppose $g(\theta, \tilde{\lambda}, \tilde{\nu}, \tilde{x}) = 0$.
Then, when all derivatives exist, the partial Jacobian of $\tilde{x}$
with respect to $\theta$ at the optimal solution $(\tilde{\lambda}, \tilde{\nu}, \tilde{x})$, namely
$J_{\theta}\tilde{x}$, can be obtained by satisfying:
\begin{equation}
J_{\theta}\tilde{x} = - \left( J_{x} g(\theta, \tilde{\lambda}, \tilde{\nu}, \tilde{x}) \right)^{-1} J_{\theta} g(\theta, \tilde{x}, \tilde{\nu}, \tilde{\lambda}).
\label{eq:theorem}
\end{equation}

Then, applying to the KKT conditions, the (partial) Jacobian with respect to $\theta$ can be defined as:
\begin{equation}
J_{\theta} g(\theta, \tilde{\lambda}, \tilde{\nu}, \tilde{x}) =
\begin{bmatrix}
J_{\theta} \nabla_{x} L(\theta, \tilde{x}, \tilde{\nu}, \tilde{\lambda}) \\
\textbf{diag}(\tilde{\lambda}) J_{\theta} (G(\theta)x - h(\theta)) \\
J_{\theta} (A(\theta) \tilde{x} - b(\theta))
\end{bmatrix}.
\label{eq:approxJac}
\end{equation}

After obtaining the optimal $\tilde{x}$, we can derive a closed-form gradient for $\theta$, enabling efficient backpropagation without altering the optimization path.

\subsection{Distillation Loss}
Assume that each $\theta_{i \rightarrow j}$ is a teacher transmitted from $j$-th neighbor silo. 
The distillation loss of $i$-th silo $\mathcal{L}^i_{\rm MD}$ based on student model loss is designed as:
\begin{equation}
\begin{aligned}
\small
\mathcal{L}^i_{\rm MD} = \beta T^2 \sum^{\mathcal{N}(i)}_{j=1}
\left( \mathcal{L}_{c}(Q^\tau_{S_i}, Q^\tau_{T_{i\rightarrow j}}) \right) + (1-\beta)\mathcal{L}_{c}(Q_{S_i},y^i_{true})
\end{aligned}
\label{eq:distil_loss}
\end{equation}
where $Q_S$ is the standard softmax output of the local student; $y^i_{true}$ is the ground-truth labels;
$\beta$ is a  hyper-parameter for controlling the importance of each loss component; $Q^\tau_{S_i}, Q^\tau_{T_{i\rightarrow j}}$ are the softened outputs of the  $i$-th local student and the $j$-th overseas teachers using the same temperature parameter $T$~\cite{hinton2015distilling}, which are computed as follows:
\begin{equation}
Q^\tau_k = \frac{\exp(l_k/T)}{\sum_{k} \exp(l_k/T)}
\label{eq:convert_softmax_loss}
\end{equation}
where the logit $l$ is outputted from the pre-final layers for both teacher and student models. Besides, as stated in Equation~\ref{eq:cross_learn}, the objective function computed for each 
$j$-th contributed transferrable weights is controlled by the corresponding EMD to ensure the learning convergence.

When the training in all silos is completed in each communication round, local model weights in all silos are aggregated to obtain global weights $\Theta = \sum^{N-1}_{i = 0 }\vartheta_i{\theta}_i$, which are further utilized for downstream fine-tuning. Note that $\vartheta \in \{0,1\}$ indicates accumulation status.


\label{subsec:EMDOpt}






\section{Experiments}
\label{Sec:Exp}

\subsection{Data Preparation}

\begin{figure}[h]
  \centering
\setlength{\tabcolsep}{2pt}
\begin{tabular}{ccc}
\shortstack{\includegraphics[width=0.45\linewidth]{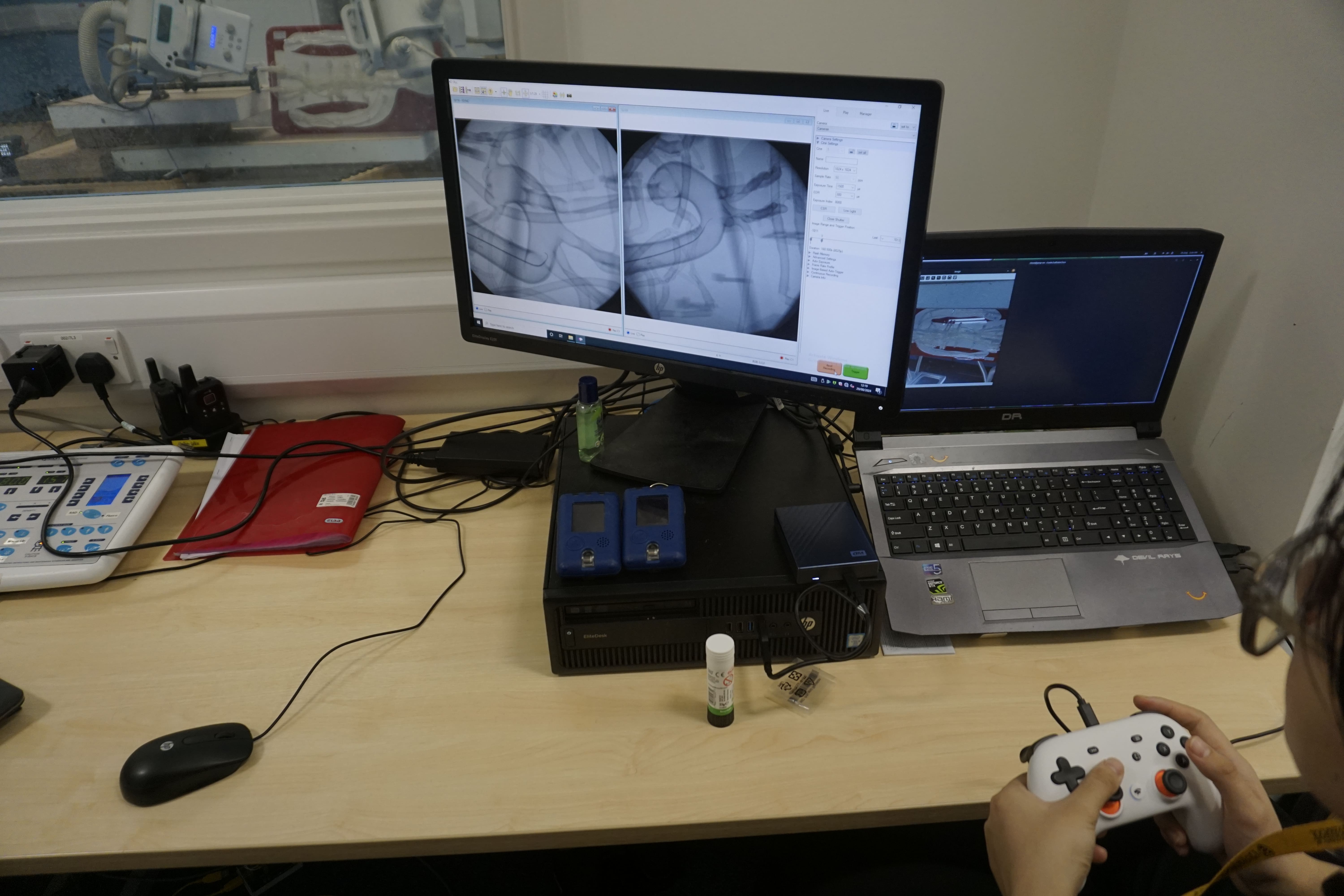}\\\footnotesize (a) Controller}&
\shortstack{\includegraphics[width=0.45\linewidth]{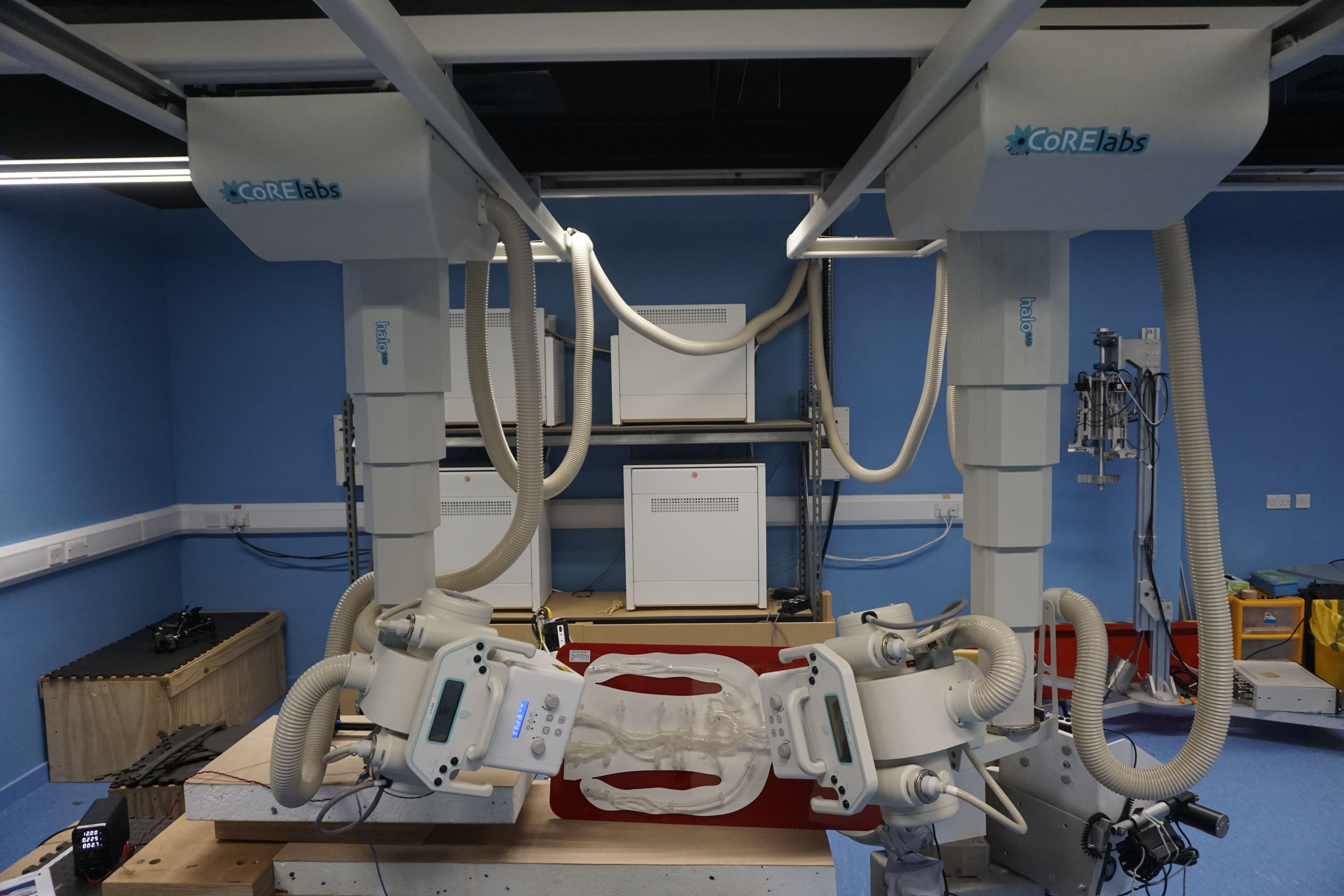}\\\footnotesize (b) Robotic Platform}\\
\end{tabular}
\vspace{-1ex}
    \caption{Data collection with endovascular robot.
    }
    \label{fig:Devices}
\end{figure}

\textbf{Robotic Setup.}
To collect large-scale X-ray images, we employ a robotic platform and a full-size silicon phantom. A surgeon uses a master device joystick to control a follower robot for cannulating three arteries: the left subclavian (LSA), left common carotid (LCCA), and right common carotid (RCCA). Fig.~\ref{fig:Devices} shows an overview of our robotics setup. During each catheterization procedure, the surgeon activates the X-ray fluoroscopy using a pedal in the operating room. The experiments are conducted using the Epsilon X-ray Generator. We develop a real-time image grabber to transmit the video feed of the surgical scene to a workstation, a computer-based device equipped with an 8-Core ARM v8.2 64-bit CPU. Overall, we collect and label 4,700 new X-ray images to create our EIPhantom dataset.

\textbf{Simulation Data.} Apart from X-ray images collected from our real robot, we also collect an EISimulation dataset from the CathSim simulator~\cite{jianu2022cathsim} for simulated X-ray images. We manually label both data from the robot and CathSim simulator to use them in downstream tasks. We note that the datasets used to train the foundation model are not being used in downstream endovascular understanding tasks. 

\begin{table}[h]
    \centering
\begin{tabular}{c|l|r}
\hline
\rowcolor[HTML]{EFEFEF}\textbf{Phase} & \textbf{Dataset}    & \textbf{\#Frames} \\ \hline\hline
 \multirow{5}{*}{\begin{tabular}[c]{@{}c@{}}Federated\\ Foundation\\ Training\end{tabular}} & CathAction~\cite{huang2024cathaction}  & $500,000$  \\  
 &\cellcolor[HTML]{EFEFEF}VESSEL12~\cite{rudyanto2014comparingVES12} & \cellcolor[HTML]{EFEFEF}$12,892$ \\
 & Drive~\cite{staal2004ridgeDrive} & $8.028$ \\ 
 &\cellcolor[HTML]{EFEFEF}SenNet~\cite{walsh2021imaging}    & \cellcolor[HTML]{EFEFEF}$7,436$  \\
 &Medical Decathlon~\cite{antonelli2022medicalDecathlon}     & $442$ \\  \hline\hline
 & \begin{tabular}[c]{@{}c@{}}EISimulation (ours)\end{tabular}    & $1,683$  \\ 
\multirow{2}{*}{\begin{tabular}[c]{@{}c@{}}Downstream\\Fine-tuning \end{tabular}} & \cellcolor[HTML]{EFEFEF}\begin{tabular}[c]{@{}c@{}} EIPhantom (ours)\end{tabular} &\cellcolor[HTML]{EFEFEF} $4,710$ \\
 & RANZCR~\cite{hansen2021radiographic}  & $33,664$  \\ 
 & \cellcolor[HTML]{EFEFEF}CathAnimal~\cite{kongtongvattana2023shape}  & \cellcolor[HTML]{EFEFEF}$25,000$  \\ \hline
\end{tabular}
\caption{X-ray datasets used in our experiments.}
\label{tab:data}
\end{table}

\textbf{Dataset Summary.} Table~\ref{tab:data} summarises datasets related to endovascular intervention~\cite{kongtongvattana2023shape,rudyanto2014comparingVES12,staal2004ridgeDrive,walsh2021imaging,antonelli2022medicalDecathlon} we use in this paper. All datasets cover different endovascular procedures with X-ray images as the main modality. The data are collected from diverse sources, including human/animal studies, human phantoms, and simulated environments.




\subsection{Federated Endovascular Foundation Model Validation} 


\textbf{Setup.} 
We first validate our proposed method (FedEFM) and compare it with different foundation models in different learning scenarios. In particular, we consider three scenarios, including Centralized Local Learning (CLL), Client-server Federated Learning (CFL)~\cite{mcmahan2017communication}, and Decentralized Federated Learning (DFL)~\cite{marfoq2020throughput}.  We note that CLL is the traditional training scenario (i.e., no federated learning) where data are merged for local training. Multiple algorithms have been conducted for the comparison purpose, including CLIP~\cite{radford2021learningCLIP}, SAM~\cite{kirillov2023segment}, LVM-Med~\cite{nguyen2023lvm}, FedAvg~\cite{mcmahan2017communication}, MOON~\cite{li2021model}, STAR~\cite{brandes2008variants}, MATCHA~\cite{wang2019matcha}, RING~\cite{marfoq2020throughput}, and CDL~\cite{do2023addressingCDL}. We use ViT~\cite{dosovitskiy2020vit} backbone in all benchmarking algorithms and train on datasets for the training phase in Table~\ref{tab:data}. Note that
our default setup is maintained at 100\% unseen label corpus.

\textbf{Results.} Table~\ref{tab:ablFM} shows the comparison with different algorithms on multiple learning scenarios. When we train ViT in CFL and DFL setup using FedAvg and MATCHA, the accuracy is only 80.9\% and 42.4\%, respectively, reflecting the inherent challenges in federated learning. Applying our proposed FedEFM method resulted in a substantial accuracy improvement to 98.2\% and 97.5\%. These results show that our proposed method can obtain competitive results even compared with the centralized training that can gather all data and only has a minor cycle time trade-off compared with most of the federated learning methods.

\begin{table}[]
\centering
\resizebox{\textwidth}{!}{
{\renewcommand{\arraystretch}{1.05}
\begin{tabular}{c|lc|c|c}
\hline
\textbf{\begin{tabular}[c]{@{}c@{}}Learning\\ Scenario\end{tabular}} & \multicolumn{2}{c|}{\textbf{Method}} & \textbf{Accuracy} & \textbf{\begin{tabular}[c]{@{}c@{}}Avg. Cycle\\ Time (ms)\end{tabular}} \\ \hline

\multirow{3}{*}{\textbf{CLL}} & \multicolumn{2}{l|}{CLIP~\cite{radford2021learningCLIP}} & 67.5 &\_\\
&\multicolumn{2}{l|}{SAM~\cite{kirillov2023segment}} & 72.4 &\_ \\
&\multicolumn{2}{l|}{LVM-Med~\cite{nguyen2023lvm}} & 98.8 &\_\\\hline \hline 
\multirow{4}{*}{\textbf{CFL}} &\multicolumn{2}{l|}{FedAvg~\cite{mcmahan2017communication} } & 80.9 &57.7\\  
 &\multicolumn{2}{l|}{MOON~\cite{li2021model}}  & 85.2 &69.2\\
   &\multicolumn{2}{l|}{STAR~\cite{brandes2008variants}}& 82.4 &63.8\\  \cline{2-5} 
 &\multicolumn{1}{c|}{\multirow{2}{*}{\textbf{FedEFM (ours)}}} & w/o EMD & 84.7 &42.5\\ \cline{3-5} 
 & \multicolumn{1}{c|}{} & \textbf{w EMD} &\textbf{\cellcolor[HTML]{EFEFEF}98.2}  &\cellcolor[HTML]{EFEFEF}61.1\\\hline\hline 
\multirow{4}{*}{\textbf{DFL}} &\multicolumn{2}{l|}{MATCHA~\cite{wang2019matcha}} & 42.4 & 43.4\\  
  &\multicolumn{2}{l|}{RING~\cite{marfoq2020throughput}} & 52.2 &73.2\\ 
  &\multicolumn{2}{l|}{CDL~\cite{do2023addressingCDL}}& 78.5 &59.9\\ \cline{2-5} 
  &\multicolumn{1}{c|}{\multirow{2}{*}{\textbf{FedEFM (ours)}}} & w/o EMD & 72.4 &47.3\\ \cline{3-5} 
 & \multicolumn{1}{c|}{} & \textbf{w EMD} &\textbf{\cellcolor[HTML]{EFEFEF}97.5}  &\cellcolor[HTML]{EFEFEF}62.1\\ \hline
\end{tabular}
}
}
\vspace{-1ex}
\caption{Foundation model performance comparison.
}
\label{tab:ablFM}
\end{table}

\subsection{Fine-tuning Results}

\textbf{Setup.} We use ViT backbone and fine-tune it using our FedEFM and different foundation models, including, CLIP~\cite{radford2021learningCLIP}, SAM~\cite{kirillov2023segment}, and LVM-Med~\cite{nguyen2023lvm}. Note that, all models are evaluated under segmentation and classification tasks in endovascular intervention. 

\textbf{Evaluation Metrics.} We use the metrics in~\cite{nguyen2023lvm,kongtongvattana2023shape} to evaluate the performance of the trained foundation model in downstream tasks. Specifically, we use Accuracy (\%) for the classification task; 2D Dice score, mIoU, and Jaccard metric are used for the segmentation task. For the segmentation task, we compare on our collected EIPhantom, EISimulation dataset, and CathAnimal~\cite{kongtongvattana2023shape}. In the classification task, we benchmark using the RANZCR dataset~\cite{hansen2021radiographic}.

\textbf{Results.} Table~\ref{tab:compare_with_FM} shows the comparison between our method and other foundation models. This table shows that the ViT backbone under our proposed algorithm outperforms other models with a clear margin. Furthermore, models trained on medical data such as LVM-Med~\cite{nguyen2023lvm} and our FedEFM archive better results compared with models trained on non-medical data such as CLIP~\cite{radford2021learningCLIP} and SAM~\cite{kirillov2023segment}. This shows that developing a domain-specific foundation model is important in the medical domain.

\begin{table*}[h]
\centering
{\renewcommand{\arraystretch}{1.2}
\caption{Fine-tuning results of different foundation models on endovascular segmentation and classification tasks. 
}
\label{tab:compare_with_FM}
\begin{adjustbox}{width=0.9\textwidth}
\begin{tabular}{l|ccccccccc|c}
\hline
\multirow{3}{*}{\textbf{Models}} & \multicolumn{9}{c|}{\textbf{Segmentation}} & \textbf{Classification} \\ \cline{2-11} 
 & \multicolumn{3}{c|}{\textbf{EIPhantom}} & \multicolumn{3}{c|}{\textbf{CathAnimal}~\cite{kongtongvattana2023shape}} & \multicolumn{3}{c|}{\textbf{EISimulation}} & \textbf{RANZCR~\cite{hansen2021radiographic}} \\ \cline{2-11} 
 & \multicolumn{1}{c|}{\textit{Dice}} & \multicolumn{1}{c|}{\textit{mIoU}} & \multicolumn{1}{c|}{\textit{Jaccard}} & \multicolumn{1}{c|}{\textit{Dice}} & \multicolumn{1}{c|}{\textit{mIoU}} & \multicolumn{1}{c|}{\textit{Jaccard}} & \multicolumn{1}{c|}{\textit{Dice}} & \multicolumn{1}{c|}{\textit{mIoU}} & \textit{Jaccard} & \textit{Accuracy} \\ \hline
\textbf{CLIP}~\cite{radford2021learningCLIP} & \multicolumn{1}{c|}{46.7} & \multicolumn{1}{c|}{23.8} & \multicolumn{1}{c|}{43.5} & \multicolumn{1}{c|}{59.1} & \multicolumn{1}{c|}{43.5} & \multicolumn{1}{c|}{52.1} & \multicolumn{1}{c|}{52.4} & \multicolumn{1}{c|}{37.3} & 32.0 & 60.4 \\ 
\textbf{SAM}~\cite{kirillov2023segment} & \multicolumn{1}{c|}{47.3} & \multicolumn{1}{c|}{29.9} & \multicolumn{1}{c|}{50.7} & \multicolumn{1}{c|}{62.2} & \multicolumn{1}{c|}{41.1} & \multicolumn{1}{c|}{58.8} & \multicolumn{1}{c|}{77.9} & \multicolumn{1}{c|}{30.5} & 51.1 & 55.4 \\ 
\textbf{LVM-Med}~\cite{nguyen2023lvm} & \multicolumn{1}{c|}{56.2} & \multicolumn{1}{c|}{31.8} & \multicolumn{1}{c|}{51.5} & \multicolumn{1}{c|}{66.6} & \multicolumn{1}{c|}{52.5} & \multicolumn{1}{c|}{70.7} & \multicolumn{1}{c|}{70.9} & \multicolumn{1}{c|}{49.1} & 61.2 & 62.3 \\ 
\rowcolor[HTML]{EFEFEF}\textbf{FedEFM (ours)} & \multicolumn{1}{c|}{\textbf{63.1}} & \multicolumn{1}{c|}{\textbf{35.5}} & \multicolumn{1}{c|}{\textbf{57.1}} & \multicolumn{1}{c|}{\textbf{67.2}} & \multicolumn{1}{c|}{\textbf{50.1}} & \multicolumn{1}{c|}{\textbf{71.8}} & \multicolumn{1}{c|}{\textbf{82.9}} & \multicolumn{1}{c|}{\textbf{63.2}} & \textbf{81.2} & \textbf{67.9} \\ \hline
\end{tabular}
\end{adjustbox}
}
\end{table*}


\begin{table*}[!t]
\centering
{\renewcommand{\arraystretch}{1.2}
\caption{Performance of different backbones when using our FedEFM for fine-tuning.
\vspace{-2ex}
}
\begin{adjustbox}{width=1\textwidth}
\begin{tabular}{l|c|ccccccccc|c}
\hline
\multirow{3}{*}{\textbf{Backbones}} & \multirow{3}{*}{\textbf{Initialize}} & \multicolumn{9}{c|}{\textbf{Segmentation}} & \textbf{Classification} \\ \cline{3-12} 
 &  & \multicolumn{3}{c|}{\textbf{EIPhantom}} & \multicolumn{3}{c|}{\textbf{CathAnimal}~\cite{kongtongvattana2023shape}} & \multicolumn{3}{c|}{\textbf{EISimulation}} & \textbf{RANZCR~\cite{hansen2021radiographic}} \\ \cline{3-12} 
 &  & \multicolumn{1}{c|}{\textit{Dice}} & \multicolumn{1}{c|}{\textit{mIoU}} & \multicolumn{1}{c|}{\textit{Jaccard}} & \multicolumn{1}{c|}{\textit{Dice}} & \multicolumn{1}{c|}{\textit{mIoU}} & \multicolumn{1}{c|}{\textit{Jaccard}} & \multicolumn{1}{c|}{\textit{Dice}} & \multicolumn{1}{c|}{\textit{mIoU}} & \textit{Jaccard} & \textit{Accuracy} \\ \hline
\multirow{2}{*}{\textbf{U-Net}~\cite{ronneberger2015u}} & From-scratch & \multicolumn{1}{c|}{48.1} & \multicolumn{1}{c|}{20.2} & \multicolumn{1}{c|}{50.2} & \multicolumn{1}{c|}{52.5} & \multicolumn{1}{c|}{42.7} & \multicolumn{1}{c|}{59.4} & \multicolumn{1}{c|}{51.1} & \multicolumn{1}{c|}{22.5} & 66.6 & 49.4 \\ 
 & \cellcolor[HTML]{EFEFEF}Fine-tuned & \multicolumn{1}{c|}{\cellcolor[HTML]{EFEFEF}\textbf{52.1}} & \multicolumn{1}{c|}{\cellcolor[HTML]{EFEFEF}\textbf{30.5}} & \multicolumn{1}{c|}{\cellcolor[HTML]{EFEFEF}\textbf{51.7}} & \multicolumn{1}{c|}{\cellcolor[HTML]{EFEFEF}\textbf{66.9}} & \multicolumn{1}{c|}{\cellcolor[HTML]{EFEFEF}\textbf{48.3}} & \multicolumn{1}{c|}{\cellcolor[HTML]{EFEFEF}\textbf{65.4}} & \multicolumn{1}{c|}{\cellcolor[HTML]{EFEFEF}\textbf{56.4}} & \multicolumn{1}{c|}{\cellcolor[HTML]{EFEFEF}\textbf{27.9}} & \cellcolor[HTML]{EFEFEF}\textbf{72.9} & \cellcolor[HTML]{EFEFEF}\textbf{56.0} \\ \hline
\multirow{2}{*}{\textbf{TransUnet}~\cite{chen2021transunet}} & From-scratch & \multicolumn{1}{c|}{46.7} & \multicolumn{1}{c|}{30.1} & \multicolumn{1}{c|}{49.9} & \multicolumn{1}{c|}{51.2} & \multicolumn{1}{c|}{44.4} & \multicolumn{1}{c|}{59.5} & \multicolumn{1}{c|}{62.2} & \multicolumn{1}{c|}{19.7} & 68.3 & 52.9 \\ 
 & \cellcolor[HTML]{EFEFEF}Fine-tuned & \multicolumn{1}{c|}{\cellcolor[HTML]{EFEFEF}\textbf{58.9}} & \multicolumn{1}{c|}{\cellcolor[HTML]{EFEFEF}\textbf{34.0}} & \multicolumn{1}{c|}{\cellcolor[HTML]{EFEFEF}\textbf{55.9}} & \multicolumn{1}{c|}{\cellcolor[HTML]{EFEFEF}\textbf{54.3}} & \multicolumn{1}{c|}{\cellcolor[HTML]{EFEFEF}\textbf{46.2}} & \multicolumn{1}{c|}{\cellcolor[HTML]{EFEFEF}\textbf{64.4}} & \multicolumn{1}{c|}{\cellcolor[HTML]{EFEFEF}\textbf{80.2}} & \multicolumn{1}{c|}{\cellcolor[HTML]{EFEFEF}\textbf{22.3}} & \cellcolor[HTML]{EFEFEF}\textbf{72.2} & \cellcolor[HTML]{EFEFEF}\textbf{58.3} \\ \hline
\multirow{2}{*}{\textbf{SwinUnet}~\cite{cao2022swin}} & From-scratch & \multicolumn{1}{c|}{47.3} & \multicolumn{1}{c|}{32.2} & \multicolumn{1}{c|}{51.7} & \multicolumn{1}{c|}{50.6} & \multicolumn{1}{c|}{43.4} & \multicolumn{1}{c|}{58.5} & \multicolumn{1}{c|}{60.8} & \multicolumn{1}{c|}{19.1} & 67.2 & 55.7 \\ 
 & \cellcolor[HTML]{EFEFEF}Fine-tuned & \multicolumn{1}{c|}{\cellcolor[HTML]{EFEFEF}\textbf{58.5}} & \multicolumn{1}{c|}{\cellcolor[HTML]{EFEFEF}\textbf{34.3}} & \multicolumn{1}{c|}{\cellcolor[HTML]{EFEFEF}\textbf{56.0}} & \multicolumn{1}{c|}{\cellcolor[HTML]{EFEFEF}\textbf{66.2}} & \multicolumn{1}{c|}{\cellcolor[HTML]{EFEFEF}\textbf{48.4}} & \multicolumn{1}{c|}{\cellcolor[HTML]{EFEFEF}\textbf{65.5}} & \multicolumn{1}{c|}{\cellcolor[HTML]{EFEFEF}\textbf{76.8}} & \multicolumn{1}{c|}{\cellcolor[HTML]{EFEFEF}\textbf{19.0}} &\cellcolor[HTML]{EFEFEF} \textbf{68.9} & \cellcolor[HTML]{EFEFEF}\textbf{62.5} \\ \hline
\multirow{2}{*}{\textbf{ViT}~\cite{dosovitskiy2020vit}} & From-scratch & \multicolumn{1}{c|}{50.9} & \multicolumn{1}{c|}{30.2} & \multicolumn{1}{c|}{50.8} & \multicolumn{1}{c|}{59.4} & \multicolumn{1}{c|}{44.7} & \multicolumn{1}{c|}{60.0} & \multicolumn{1}{c|}{72.1} & \multicolumn{1}{c|}{61.4} & 74.5 & 60.6 \\ 
 &\cellcolor[HTML]{EFEFEF} Fine-tuned & \multicolumn{1}{c|}{\cellcolor[HTML]{EFEFEF}\textbf{63.1}} & \multicolumn{1}{c|}{\cellcolor[HTML]{EFEFEF}\textbf{35.5}} & \multicolumn{1}{c|}{\cellcolor[HTML]{EFEFEF}\textbf{57.1}} & \multicolumn{1}{c|}{\cellcolor[HTML]{EFEFEF}\textbf{67.2}} & \multicolumn{1}{c|}{\cellcolor[HTML]{EFEFEF}\textbf{50.1}} & \multicolumn{1}{c|}{\cellcolor[HTML]{EFEFEF}\textbf{71.8}} & \multicolumn{1}{c|}{\cellcolor[HTML]{EFEFEF}\textbf{82.9}} & \multicolumn{1}{c|}{\cellcolor[HTML]{EFEFEF}\textbf{63.2}} & \cellcolor[HTML]{EFEFEF}\textbf{81.2} & \cellcolor[HTML]{EFEFEF}\textbf{67.9} \\ \hline
\end{tabular}
\end{adjustbox}
\label{tab:backbone}
}
\end{table*}

\subsection{Abliation Study}

\begin{figure}[t]
  \centering
  \includegraphics[width=0.7\linewidth, height=0.7\linewidth]{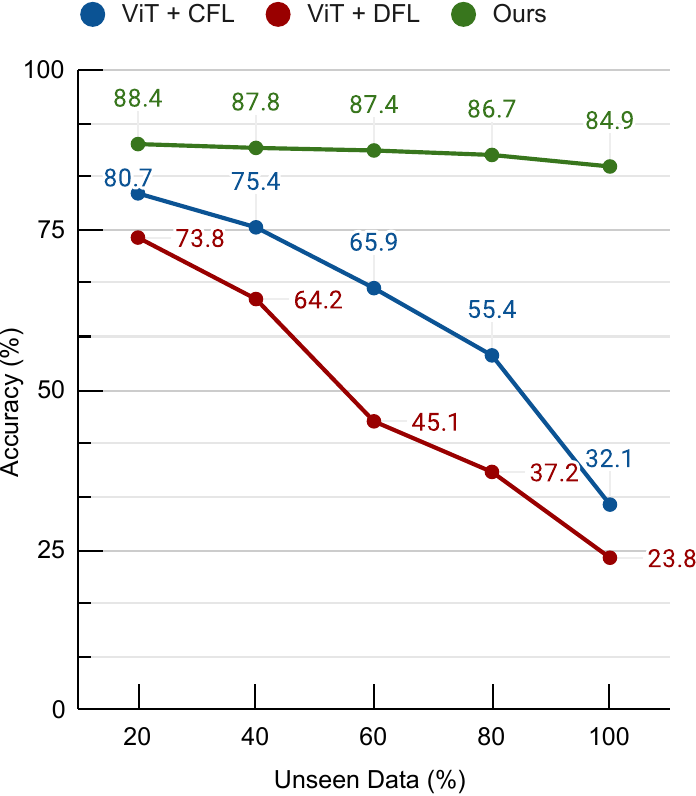}
  \vspace{-1ex}
 \caption{Results with different unseen data proportions. \vspace{-3ex}
 }
 \label{fig:UnseenAnalysis}
\end{figure}

\textbf{Unseen Data Proportion Analysis.} Fig.~\ref{fig:UnseenAnalysis} presents an analysis of our method under different percentages of unseen data. In this experiment, we assume that each silo (hospital) only keeps an amount of data (e.g., human/animal/simulated X-ray) where their data corpus only shares the similarity in a given percentage. A 100\% unseen data corpus means that the data of each hospital silo have no similarity in their data types compared to others. As the percentage of unseen data types increases, we observe a notable decline in the accuracy of the baseline on CFL and DFL scenarios. However, our proposed approach demonstrates remarkable resilience to unseen data, maintaining high accuracy even when confronted with a higher percentage of unfamiliar semantic data. In specific instances, when all data labels are unseen (100\%), ViT under CFL and DFL scenarios exhibit significantly lower accuracies at 32.1\% and 23.8\%, respectively. In contrast, our approach achieves an accuracy of 84.9\%, showcasing its effectiveness in handling unseen data. 

\textbf{Backbones Analysis.}
We verify the stability of our method on different networks, including UNet~\cite{ronneberger2015u}, TransUNet~\cite{chen2021transunet}, and SwinUnet~\cite{cao2022swin}, and ViT~\cite{dosovitskiy2020vit} under federated learning scenario. Table~\ref{tab:backbone} shows the performance of the different backbones when we fine-tune them using our FedEFM. Table~\ref{tab:backbone} demonstrates that using our foundation model to initialize the weights of those backbones significantly improves the results. These results validate the effectiveness of our training process in addressing the unseen data problem, and our FedEFM is useful for different backbones in endovascular downstream tasks.

\begin{figure}[!t] 
  \centering
  \Large
\resizebox{\linewidth}{!}{
\setlength{\tabcolsep}{2pt}
\begin{tabular}{ccccccc}
\Large
\rotatebox[origin=l]{90}{\hspace{0.35cm} Animal}&
\shortstack{\includegraphics[width=0.33\linewidth]{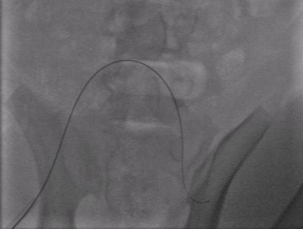}}&
\shortstack{\includegraphics[width=0.33\linewidth]{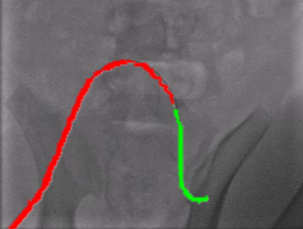}}&
\shortstack{\includegraphics[width=0.33\linewidth]{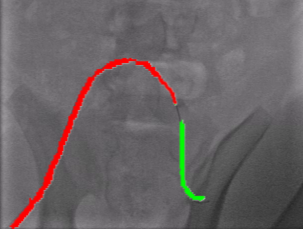}}&
\shortstack{\includegraphics[width=0.33\linewidth]{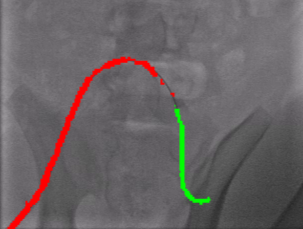}}&
\shortstack{\includegraphics[width=0.33\linewidth]{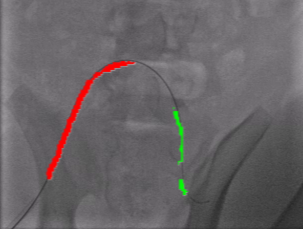}}&
\shortstack{\includegraphics[width=0.33\linewidth]{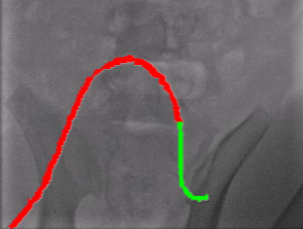}}\\[3pt]
\rotatebox[origin=l]{90}{\hspace{0.05cm} Phantom}&
\shortstack{\includegraphics[width=0.33\linewidth]{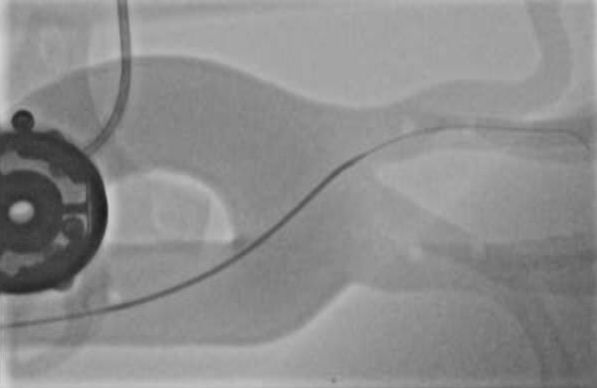}}&
\shortstack{\includegraphics[width=0.33\linewidth]{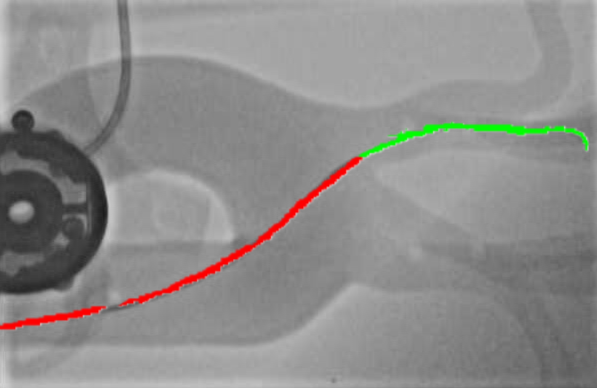}}&
\shortstack{\includegraphics[width=0.33\linewidth]{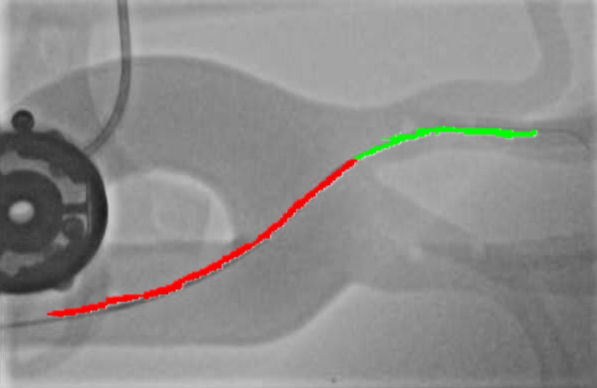}}&
\shortstack{\includegraphics[width=0.33\linewidth]{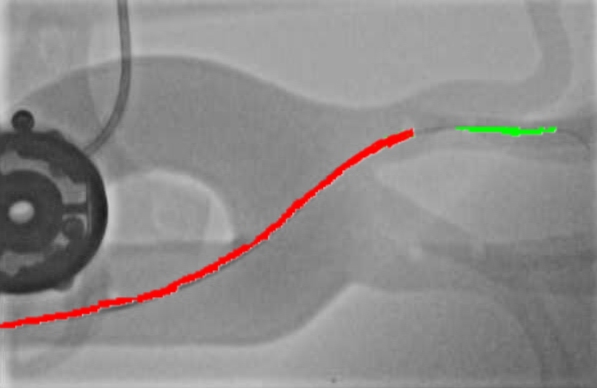}}&
\shortstack{\includegraphics[width=0.33\linewidth]{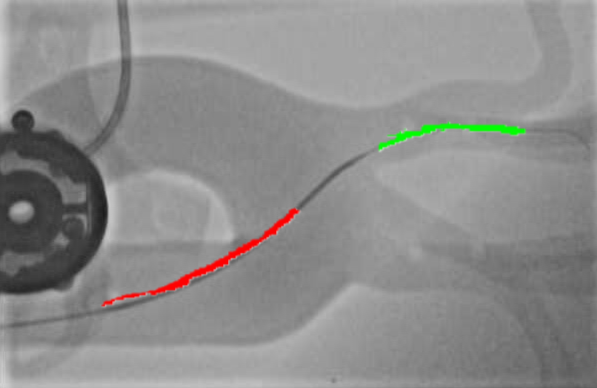}}&
\shortstack{\includegraphics[width=0.33\linewidth]{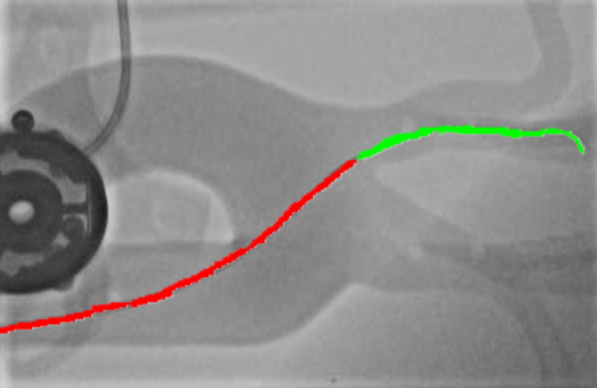}}\\[3pt]
\rotatebox[origin=l]{90}{\hspace{0.1cm}Simulation}&\shortstack{\includegraphics[width=0.33\linewidth]{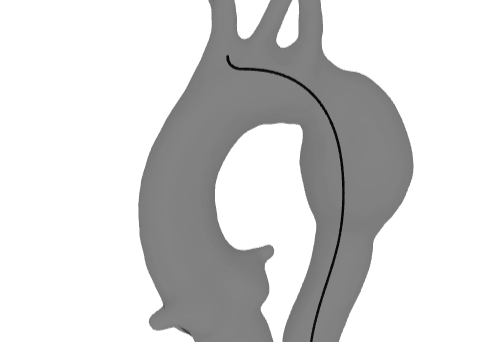} \\  Input}&
\shortstack{\includegraphics[width=0.33\linewidth]{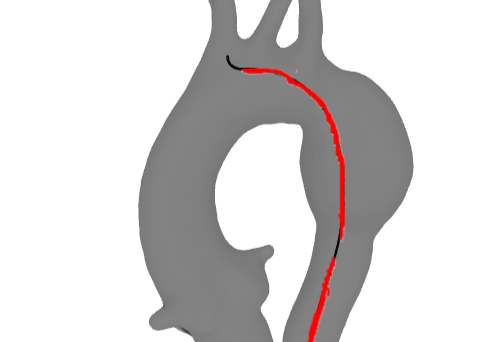} \\ Ground Truth}&
\shortstack{\includegraphics[width=0.33\linewidth]{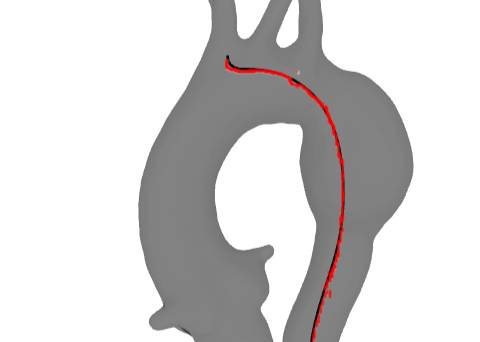} \\ LVM-Med}&
\shortstack{\includegraphics[width=0.33\linewidth]{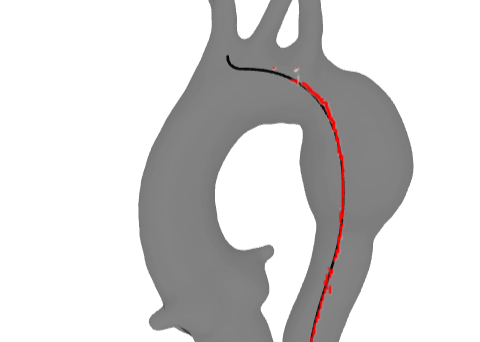} \\ SAM}&
\shortstack{\includegraphics[width=0.33\linewidth]{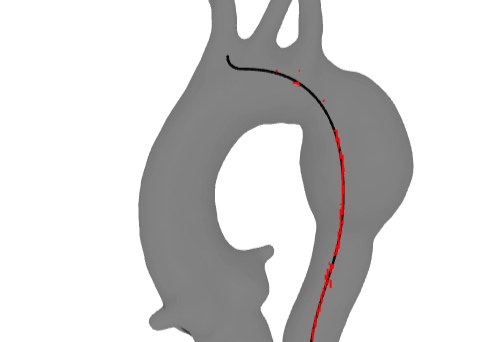} \\ CLIP}&
\shortstack{\includegraphics[width=0.33\linewidth]{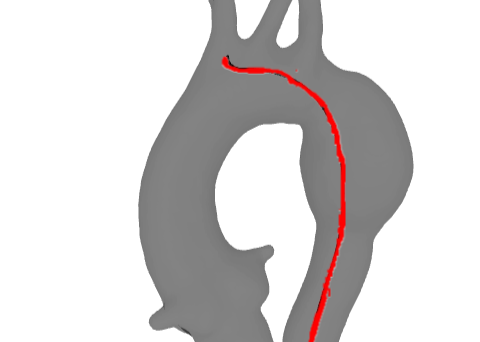} \\ Ours}\\ [1pt]
\end{tabular}
}
    \caption{Catheter and guidewire segmentation between methods. Red lines are catheters and green ones are guidewires.} 
    \label{fig:SegVis}
\end{figure}

\textbf{Qualitative Results.} Fig.~\ref{fig:SegVis} illustrates the catheter and guidewire segmentation results of fine-tuning ViT on our method and different foundation models. The visualization portrays that our method excels in accurately delineating the catheter and guidewire structures, showcasing superior segmentation performance compared to other approaches. This figure further confirms that we can successfully train a federated endovascular foundation model without collecting users' data and the trained foundation model is useful for the downstream segmentation task.

\subsection{Limitations} While our proposed approach demonstrates significant potential, it is subject to certain limitations that warrant further investigation. Firstly, the requirement for additional weight exchange among silos extends the overall training time. However, this limitation is mitigated to some extent by the higher convergence speed of our method compared to other approaches. Additionally, our method is designed for deployment in silos with strong GPU computing resources, but the varying hardware capabilities present in many real-world federated learning networks necessitate further examination. Overcoming these limitations will open new research in federated foundation learning for endovascular interventions and other medical applications. Furthermore, addressing the challenges of managing heterogeneous data distributions and ensuring robust data privacy remains a critical focus. Moving forward, we plan to extend our approach to robotic-assisted endovascular surgery and other areas, such as pathology, to further investigate the application of federated foundation models in medical imaging and robotic systems.


\section{Conclusion}
\label{Sec:Conclusion}

We present a new approach to train an endovascular foundation model in a federated learning setting, leveraging differentiable Earth Mover's Distance and knowledge distillation to handle the unseen data issue. Our method ensures that once the foundational model is trained, its weights can be effectively fine-tuned for downstream tasks, thereby enhancing performance. Our approach achieves state-of-the-art results and contributes to the field of endovascular intervention, particularly by addressing the critical issue of data sharing in the medical domain. By enabling weight exchange among local silos and fostering knowledge transfer, our method improves model generalization while preserving data privacy.
Experimental results across various endovascular imaging tasks validate the efficacy of our approach, demonstrating its potential for application in privacy-sensitive medical domains. We will release our implementation and trained models to facilitate reproducibility and further research.


\bibliographystyle{IEEEtran}
\bibliography{IEEEfull}

\end{document}